\documentclass[letterpaper, 10 pt, conference]{ieeeconf}  % Comment this line out if you need a4paper

\IEEEoverridecommandlockouts                              % This command is only needed if 
\usepackage{amsmath} % assumes amsmath package installed
\usepackage{todonotes}
\usepackage{url}
\usepackage{makecell}
\usepackage{algorithm}
\usepackage{algorithmic}
\usepackage{footnote}
\usepackage{scrextend}
\usepackage{subcaption}
\usepackage{multirow}
\usepackage{pifont}
\definecolor{myblue}{rgb}{0.1,0.1,0.5}
\definecolor{correct_g}{rgb}{0, 0.8, 0}
\definecolor{fail_r}{rgb}{0.82, 0.06, 0.12}
\definecolor{grayfig}{HTML}{FF8000}
\definecolor{bluefig}{HTML}{007FFF}
\usepackage{hyperref}
\usepackage[capitalise]{cleveref}
\usepackage{amssymb}
\usepackage[keeplastbox]{flushend}
\usepackage{tikz, pgfplots}
\pgfplotsset{compat=1.16}
\pgfplotsset{every tick label/.append style={font=\small}}
\usepackage{etoolbox}\AtBeginEnvironment{algorithmic}{\small}

\usepackage[style=ieee, url=false, doi=false, natbib=true, mincitenames=1, maxcitenames=1]{biblatex}

\title{\LARGE \bf
Out-of-Distribution Detection for Automotive Perception
}

\author{Julia Nitsch$^{1,2}$, Masha Itkina$^{3}$, Ransalu Senanayake$^{3}$, Juan Nieto$^{2}$, Max Schmidt$^{1}$, Roland Siegwart$^{2}$,\\ Mykel J. Kochenderfer$^{3}$, and Cesar Cadena$^{2}$% <-this % stops a space
\thanks{$^{1}$Ibeo Automotive Systems GmbH, 
        {\tt\small \{julia.nitsch, max.schmidt\}@ibeo-as.com}}%
\thanks{$^{2}$Autonomous Systems Lab, ETH Zurich,
        {\tt\small \{jnieto, rsiegwart, cesarc\}@ethz.ch}}%
\thanks{$^{3}$Stanford Intelligent Systems Laboratory, Stanford University,
        {\tt\small \{mitkina, ransalu, mykel\}@stanford.edu}}%       
}

\begin{document}

\setlength{\abovedisplayskip}{3pt}
\setlength{\belowdisplayskip}{3pt}

\maketitle
\thispagestyle{empty}
\pagestyle{empty}

%%%%%%%%%%%%%%%%%%%%%%%%%%%%%%%%%%%%%%%%%%%%%%%%%%%%%%%%%%%%%%%%%%%%%%%%%%%%%%%%
\begin{abstract}
Neural networks (NNs) are widely used for object classification in autonomous driving. However, NNs can fail on input data not well represented by the training dataset, known as out-of-distribution (OOD) data. A mechanism to detect OOD samples is important for safety-critical applications, such as automotive perception, to trigger a safe fallback mode. NNs often rely on softmax normalization for confidence estimation, which can lead to high confidences being assigned to OOD samples, thus hindering the detection of failures.
This paper presents a method for determining whether inputs are OOD, which does not require OOD data during training and does not increase the computational cost of inference. The latter property is especially important in automotive applications with limited computational resources and real-time constraints. Our proposed approach outperforms state-of-the-art methods on real-world automotive datasets.
\end{abstract}
%%%%%%%%%%%%%%%%%%%%%%%%%%%%%%%%%%%%%%%%%%%%%%%%%%%%%%%%%%%%%%%%%%%%%%%%%%%%%%%%
\section{Introduction}
\begin{figure}
    \centering
    \includegraphics[width=0.7\linewidth]{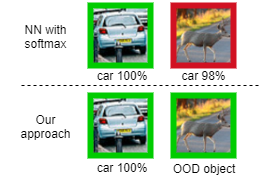}
    \caption{A NN-based classifier with a softmax output layer trained on road users can reliably classify cars that were not present in $D_{in}$. However, if an OOD object, such as a deer, is processed during inference, the NN may incorrectly assign high confidence to it. Our approach can detect these OOD objects and classify them as such while continuing to reliably classify $D_{in}$ objects.}
    \label{fig:teaser}
\vspace{-7mm}
\end{figure}

Neural network (NN) perception systems currently provide state-of-the-art object classification performance~\cite{qi2018frustum, bhattacharyya2020deformable, zhu2019class, zhu2020ssn}. Softmax normalization is ubiquitously used by NNs for uncertainty estimation. However, such NNs tend to fail on out-of-distribution (OOD) data by assigning incorrect classifications with high confidence~\cite{sunderhauf2018limits}. Although softmax tends to perform well on data that follows the training distribution ($D_{in}$), it assigns overconfident values to OOD samples ($D_{out}$) due to the fast-growing exponential function~\cite{hendrycks2016baseline}. If novel objects appear during inference, the NN could classify them incorrectly with high confidence~\cite{hendrycks2016baseline, nguyen2015posterior, yu2011calibration, provost1998case, nguyen2015deep}. For example, if a NN is designed to classify road users and is therefore trained on samples of pedestrians, cars, trucks, and cyclists, it may eventually classify a deer with high confidence as a car. The deer is an OOD sample for the NN since it was not included in the training data and, thus, the NN misclassifies it as shown in \cref{fig:teaser}. Hence, these confidences output by the NN pose multiple risks such as decreasing the overall perception accuracy and hiding failures, which reduces system reliability and forces autonomous vehicles to take more conservative actions. To avoid the restriction to conservative actions in general, it is important to detect these perception errors and only then take appropriate measures such as switching to a safe fallback mode in order to ensure safe execution~\cite{sunderhauf2018limits}.

The literature discusses various approaches for identifying $D_{out}$. One category of approaches uses NN architectures that have implicit OOD detection~\cite{ovadia2019can, sedlmeier2019uncertainty, loquercio2020general}, but these methods often increase computational costs. Another category of approaches is dedicated OOD detection methods, which include training techniques such as auxiliary losses or modifications to the NN architecture~\cite{hendrycks2016baseline, hendrycks2018deep, devries2018learning}. These methods usually require OOD data during development. This paper presents a threshold-based OOD detection approach that combines auxiliary training techniques with post hoc statistics that capture network characteristics after training. This combination improves the area under the precision recall curve (AUPR) metric compared to state-of-the-art methods. Additionally, our approach does not require $D_{out}$ during training and comes without additional computational costs at inference time. The latter attribute, in particular, is important for automotive applications that run on limited resources with real-time constraints. During training, our proposed method uses an auxiliary Generative Adversarial Network (GAN)~\cite{goodfellow2014generativ} to produce $D_{out}$ data that encourages the object classifier to assign low confidences to samples on or outside the decision boundary. Post hoc, the parameters of class-conditioned Gaussian distributions over the bottleneck layer of the NN are computed from the training data. A sample is classified as $D_{out}$ during inference based on its distance to the previously computed Gaussian distributions. Our proposed method is validated on an object classification task using real-world automotive datasets, including KITTI~\cite{Geiger2012CVPR} and nuScenes~\cite{nuscenes2019ceasar}. 

Our contributions are as follows. We present an OOD detection approach that does not require an externally collected $D_{out}$ dataset and comes without additional computational costs during inference. We empirically validate our OOD detection method on real-world data with multiple post hoc distance metrics, demonstrating superior performance to state-of-the-art approaches.

\section{Related Work}
OOD detection can be categorized into implicit (\cref{sec:related_work:ood_dist}) and explicit (\cref{sec:related_work:dedicated_ood}) detection techniques. Implicit OOD detection techniques can naturally detect OOD samples as a byproduct of their design, while dedicated OOD detection approaches consider auxiliary losses during training, additional architectural components, or post hoc network statistics.

\subsection{Implicit OOD Detection Techniques}
\label{sec:related_work:ood_dist}
Bayesian Neural Networks (BNNs) can identify samples from $D_{out}$ by design unlike Maximum Likelihood Estimation (MLE)-based approaches~\cite{ovadia2019can, jospin2020hands}. BNNs represent model uncertainty using parameterized distributions over NN weights. NN ensembles can also represent model uncertainty by using a collection of weights~\cite{jospin2020hands, wehenkel2020you}. Both of these methods can be used for OOD detection. Although BNNs and NN ensembles are promising research directions, current state-of-the-art automotive perception usually relies on MLE-based deep NNs~\cite{qi2018frustum, bhattacharyya2020deformable, zhu2019class, zhu2020ssn} due to lower computational costs. Hence, in our paper, we follow this line of research.
Monte Carlo (MC) dropout~\cite{gal2016dropout} has gained popularity for uncertainty-based OOD detection~\cite{sedlmeier2019uncertainty, loquercio2020general}. MC dropout is a simple, effective method for epistemic uncertainty estimation that does not require an explicit $D_{out}$ dataset to be available during training. In contrast to BNNs, which are trained to compute model uncertainty, MC dropout measures the uncertainty by applying dropout to the NN during inference. For this, MC dropout requires multiple forward passes during inference, which is unsuitable for applications that run on limited resources and have real-time constraints.

Normalizing flows (NFs) have also been used for OOD detection using the learned feature space~\cite{marchal2020learning}. However, the memory footprint of NFs is unsuitable for larger inputs and, thus, \citet{wellhausen2020safe} apply NFs only to the bottleneck vector of a NN. We follow the latter approach in modeling distributions over the bottleneck vector, but instead of NFs, we fit class-conditioned Gaussian distributions following~\citet{lee2018simple}, which add no extra memory footprint to the NN.

\subsection{Dedicated OOD Detection Techniques}
\label{sec:related_work:dedicated_ood}
Variational Autoencoders (VAEs)~\cite{an2015variational, lee2020detect, sensoy2020uncertainty} and adversarially trained AEs~\cite{sabokrou2018adversarially} can specifically be developed for OOD detection. Latent space distances and reconstruction losses of bidirectional GANs~\cite{donahue2016bigan} are used for $D_{out}$ classification~\cite{zenati2018efficient, zenati2018adversarially}. However, these methods incur additional computational costs during inference.

NN architecture components like decoders~\cite{hendrycks2016baseline}, additional output neurons~\cite{devries2018learning}, or auxiliary loss functions~\cite{hendrycks2018deep} are used for OOD detection. However, these methods require explicit $D_{out}$ datasets during training. \citet{golan2018deep} circumvent the need for an explicit $D_{out}$ dataset by applying geometric transformations to the input data, such as flipping, rotating, or translating images, during training. However, the same transformation must be applied during inference to detect OOD samples, which adds computational cost. Classifiers have been trained on feature space distances~\cite{snell2017prototypical, tan2019domain} to detect OOD samples during inference. We do not train explicitly on feature distances but apply an auxiliary loss to assign low confidences to generated OOD examples. We thus maximize the distance from our generated $D_{out}$ samples to the $D_{in}$ dataset implicitly. 

Post hoc methods, such as that proposed by \citet{lee2018simple}, compute network statistics on an already trained network to detect OOD samples. Combining auxiliary training methods from \citet{hendrycks2018deep} with post hoc statistics from \citet{lee2018simple} has been proposed by \citet{papadopoulos2019outlier}, but they require a carefully designed $D_{out}$ dataset. \citet{lee2017training} avoid the design of $D_{out}$ by using an auxiliary GAN to generate $D_{out}$ during training and an additional loss function that assigns uniform distribution to $D_{out}$. 
In our work, we also use a GAN to produce $D_{out}$ samples during training. Following \citet{papadopoulos2019outlier}, we combine this auxiliary training procedure with post hoc statistics and investigate multiple distance metrics for OOD detection. With this combination, we circumvent the need for $D_{out}$ dataset collection and outperform the state-of-the-art approach from \citet{lee2017training}.

\section{Method}
\begin{figure}
    \centering
    \includegraphics[width=\linewidth]{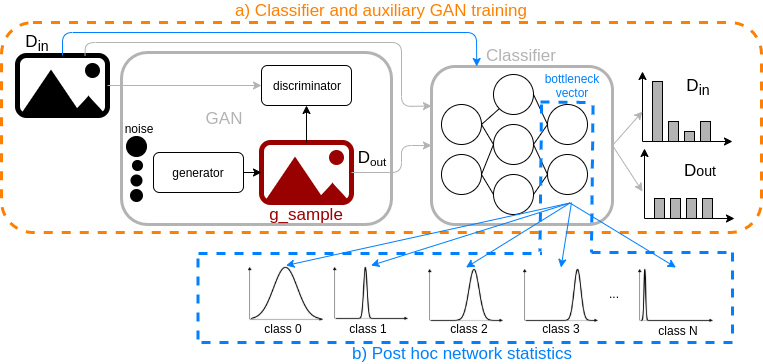}
    \caption{\textcolor{grayfig}{a)} depicts the processing of $D_{in}$ and the generation of $D_{out}$ on the decision boundary of the classifier. \textcolor{bluefig}{b)} depicts the post hoc computation of class-conditioned Gaussian distributions on $D_{in}$. During inference, a threshold identifies $D_{out}$ based on the distance to these distributions.}
    \label{fig:architecture_training}
\vspace{-7mm}
\end{figure}
We propose an uncertainty-based OOD detection approach that combines auxiliary training techniques with post hoc statistics. An overview of the proposed method is shown in \cref{fig:architecture_training}. Similar to \citet{lee2017training}, we use an auxiliary GAN to produce $D_{out}$ during training to encourage the object classifier to assign low confidence to samples on and outside the decision boundary. We extend the approach of \citet{lee2017training} by proposing an additional post hoc component to the OOD detection method, thereby improving performance. We compute the parameters of class-conditioned Gaussian distributions over the weights of the bottleneck layer of the NN based on the training data. Class-conditioned Gaussian distributions are a reasonable choice for modeling distributions on pre-trained features according to the analysis by~\citet{lee2018simple}. During inference, a distance computation to the Gaussian distributions facilitates the identification of $D_{out}$ through an empirically determined threshold. We explore different distance metrics in our evaluation.
Our proposed approach does not require an explicit OOD dataset during training and does not add computation costs during inference which makes it suitable for real-world applications.

\begin{figure}
\centering
\begin{subfigure}[t]{0.3\textwidth}
    \centering
    \includegraphics[width=\textwidth]{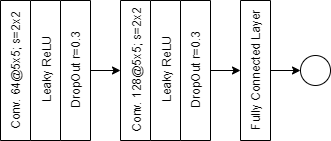}
    \caption{Discriminator Architecture}
\label{fig:architecture_disc}
\end{subfigure}
\par\medskip 
\begin{subfigure}[t]{0.4\textwidth}
\centering
    \includegraphics[width=\textwidth]{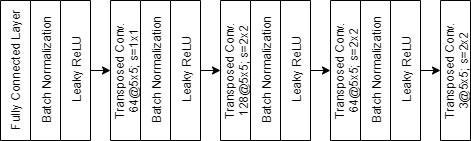}
    \caption{Generator Architecture}
\label{fig:architecture_gen}
\end{subfigure}
\caption{The discriminator (\cref{fig:architecture_disc}) consists of two convolutional layers with $5\times5$ kernels, strides of $2\times2$, and leaky ReLU activations followed by a dropout layer with a dropout rate of $0.3$. The first and second layers produce $64$ and $128$ feature maps, respectively. The convolutions are followed by a fully connected layer with a single output. The first block of the generator (\cref{fig:architecture_gen}) consists of a fully connected layer with batch normalization and a leaky ReLU activation. Then, the network contains three transposed convolution blocks with $5\times5$ kernels, batch normalization, and leaky ReLU activations. The last block consists of only a transposed convolution with $5\times5$ kernel.}
\vspace{-7mm}
\end{figure}

\subsection{Training}
\label{sec:method:training}
A GAN architecture consists of two components: a generator and a discriminator. The generator synthesizes data samples such that the discriminator would incorrectly classify these generated samples ($D_{out}$) as regular input data ($D_{in}$). The discriminator is optimized to distinguish between the generated samples and the real input. The discriminator loss in \cref{eq:l_d}, the generator loss in \cref{eq:l_g}, and the object classification loss in \cref{eq:l_cls} are jointly trained (see \cref{alg:joint_loss}) following~\cite{lee2017training}.

The discriminator loss is computed to distinguish samples from $D_{in}$ and $D_{out}$, and is thus defined as the sum of the binary cross-entropy (BCE) losses from $D_{in}$ and from $D_{out}$: 
\begingroup
\small
\begin{align}
    \mathcal{L}_{d}=\mathrm{BCE}\left(\mathrm{dis}(D_{in}), \mathbf{1}\right) + \mathrm{BCE}\left(\mathrm{dis}(\mathrm{gen}(\mathbf{n})), \mathbf{0}\right)
\label{eq:l_d}
\end{align}
\endgroup
where $\mathbf{n}$ is the standard Gaussian noise vector, $\mathrm{gen}$ is the generator output, and $\mathrm{dis}$ is the discriminator output.
In our construction, the generator loss in \cref{eq:l_g} features an additive term that serves to encourage the network to output samples on the decision boundary of the object classifier. Hence, the generator is not only trained to fool the discriminator, but also to encourage the object classifier to assign low confidences to samples on or outside the decision boundary. This is achieved through the additive Kullback-Leibler (KL) divergence term between the classifier output and the uniform distribution. The generator loss is: 
\begingroup
\small
\begin{align}
    \mathcal{L}_{g}=\mathrm{BCE}\left(\mathrm{dis}(\mathrm{gen}(\mathbf{n})), \mathbf{1}\right) + \text{KL}\left[\mathrm{cls}\left(\mathrm{gen}(\mathbf{n})\right)\Big\Vert \frac{\mathbf{1}}{n_c}\right]
\label{eq:l_g}
\end{align}
\endgroup
where $n_c$ is the number of classes, $\frac{\mathbf{1}}{n_c}$ is the uniform distribution over classes, and $\mathrm{cls}$ is the output of the classification network with softmax normalization. Similarly, the classifier weights are updated to classify $D_{in}$ correctly using the cross entropy loss (CE) and to assign a uniform distribution to $D_{out}$ with the same KL term as in $\mathcal{L}_{g}$ (see \cref{eq:l_g}). This KL term trains the generator to create samples on the decision boundary of the classifier and the same term encourages the classifier to assign uniform distribution to these samples. We denote the ground truth label as $\boldsymbol{\ell}$. The classifier loss is defined as: 
\begingroup
\small
\begin{align}
\begin{split}
   \mathcal{L}_{cls}=\mathrm{CE}\left(\mathrm{cls}(D_{in}), \boldsymbol{\ell}\right) +  \text{KL}\left[\mathrm{cls}\left(\mathrm{gen}(\mathbf{n}\right)) \Big\Vert \frac{\mathbf{1}}{n_c}\right].
\end{split}\label{eq:l_cls}
\end{align}
\endgroup
The sequential learning procedure is demonstrated in detail in \cref{alg:joint_loss}.
\begin{algorithm}
\caption{Joint Training Loss}
\label{alg:joint_loss}
\begin{algorithmic}
\FOR{$epoch$ \textbf{in} $epochs$}
\STATE $batches\_din \gets split\:D_{in}\:into\:batches\:of\: batch\_size$
  \FOR{($data$, $\boldsymbol{\ell}$) \textbf{in} $batches\_din$}
  \STATE $\mathbf{n} = \text{sample\_n}\left(\right)$
  
  \STATE $\text{\# 1) update discriminator}$
  \STATE $\mathcal{L}_{d} = \text{BCE}\left(\text{dis}\left(data\right), \mathbf{1}\right) + \text{BCE}\left(\text{dis}\left(\text{gen}\left(\mathbf{n}\right)\right), \mathbf{0}\right)$
  \STATE update dis w.r.t. $\mathcal{L}_{d}$.
  
  \STATE $\text{\# 2) update generator}$
  \STATE $\mathcal{L}_{g}=\text{BCE}\left(\text{dis}\left(\text{gen}\left(\mathbf{n}\right)\right), \mathbf{1}\right) + \text{KL}\left(\text{cls}\left(\text{gen}\left(\mathbf{n}\right)\right), \frac{\mathbf{1}}{\left|n_c\right|}\right)$
   \STATE update gen w.r.t. $\mathcal{L}_{gen}$.
  
  \STATE $\text{\# 3) update classifier}$
  \STATE $\mathcal{L}_{cls}=\text{CE}\left(\text{cls}\left(data\right), \boldsymbol{\ell}\right) + \text{KL}\left(\text{cls}\left(\text{gen}\left(\mathbf{n}\right)\right), \frac{\mathbf{1}}{\left|n_c\right|}\right)$
   \STATE update cls w.r.t. $\mathcal{L}_{cls}$.
  \ENDFOR
\ENDFOR
\end{algorithmic}
\end{algorithm}

In contrast to~\cite{lee2017training}, we propose a different architecture for the discriminator (see \cref{fig:architecture_disc}) and generator (see \cref{fig:architecture_gen}) due to stability issues encountered during training on automotive datasets. The main differences between our discriminator architecture and that of \citet{lee2017training} are as follows. We use a smaller number of convolutional layers with larger $5\times5$ kernels. We also replace batch normalization with dropout regularization since we empirically found better performance with this regularization method. We hypothesize that the larger kernels might be beneficial due to larger input sizes compared to the datasets used in~\cite{lee2017training}. Within the generator, we insert a fully connected layer and again use kernels of size $5\times5$. We found empirically that the addition of the fully connected layer accelerated the generator training. Similar to \citet{lee2017training}, we use the VGG-13 architecture~\cite{simonyan2014very} for the classification network. We use a 100-dimensional standard Gaussian noise vector $\mathbf{n}$ as input to the generator network to produce $D_{out}$. The discriminator ($\mathrm{dis}$) takes as input data of shape $height \times width \times 3$ with a value range of $[0, 1]$. The discriminator outputs a confidence value in $[0, 1]$, where $1$ is the highest confidence that the input belongs to $D_{in}$. The generator ($\mathrm{gen}$) processes $\mathbf{n}$ to synthesize data samples of the same shape and value range as $D_{in}$.
All weights are initialized with the Xavier initialization~\cite{glorot2010understanding}, biases are initialized with zero, and the network is trained with the Adam optimizer~\cite{kingma2014adam}. The initial learning rate of $2 \times 10^{-4}$ is exponentially decreased every $30000$ iterations with a decay rate of $0.5$. The network is trained for 100 epochs with a batch size of $128$.

\subsection{Post Hoc Network Statistics}
\label{sec:method:post_training}
Having trained the network, we now want to compute the parameters of class-conditioned Gaussian distributions over the logits, also referred to as the bottleneck vector, of the classification network. The class-conditioned Gaussian is a reasonable modeling choice for the parameter distribution according to \citet{lee2018simple}. The distance to the Gaussian distribution deemed most likely by the softmax distribution is used to identify OOD samples. We first consider the cosine similarity measure, which is a known distance metric for comparing high dimensional feature vectors and, thus, an effective means for OOD detection: 
\begingroup
\small
\begin{align}
    CosSim\left(\mathbf{x}, \boldsymbol{\mu}_c\right) = \frac{\mathbf{x} \cdot \boldsymbol{\mu}_c}{\left|\mathbf{x}\right| \; \left|\boldsymbol{\mu}_c\right|}
\end{align} 
\endgroup
where $\mathbf{x}$ is the logit vector from the classifier for a test input, $\boldsymbol{\mu}_c$ is the mean vector, and $c$ is the most likely class according to the softmax distribution. Since the cosine similarity does not take into account the dispersion of the features, we also consider computing the Mahalanobis distance as suggested by~\citet{lee2018simple}:
\begingroup
\small
\begin{align}
    d_{M_c} = \sqrt{\left(\mathbf{x} - \boldsymbol{\mu}_c\right)^T\boldsymbol{\Sigma}_c^{-1}\left(\mathbf{x} - \boldsymbol{\mu}_c\right)}
\end{align} 
\endgroup
where $\boldsymbol{\Sigma}_c$ is the covariance matrix. Since $d_{M_c}$ is not bounded, which is required for the common interpretation of confidence measures, we compute the confidence based on the chi-square ($\chi^2$) distribution. $d_{M_c}^2$ follows the $\chi^2$ distribution where the degrees of freedom correspond to the feature dimensions. This probability is then used for OOD detection~\cite{rousseeuw1990unmasking, rousseeuw1999fast}. Unlike our proposed method, \citet{lee2018simple} require an explicit, pre-determined $D_{out}$ during training in order to perform OOD detection during inference.

\section{Experiments}
We consider two classification training regimes in our experiments: (1) the classifier is trained using the standard cross entropy loss (CE loss) and (2) the classifier is trained using the sequential losses in \cref{eq:l_d} - \cref{eq:l_cls} (joint loss). For both sets of experiments, the predicted class label is the most likely class according to the softmax layer. For the predicted class, the following post hoc confidence measures are computed to detect $D_{out}$: largest softmax probability mass amongst all classes (softmax), mutual information (MI) from MC dropout~\cite{gal2016dropout}, cumulative density function (CDF) of the $\chi^2$ distribution (ours), and cosine similarity (ours).
The joint loss training with softmax probabilities inspired by~\cite{lee2017training}, serves as a baseline to our proposed combination with post hoc statistics ($\chi^2$ and CosSim). We compare our proposed method against MC dropout using $T=100$ inference steps as it is a popular method for OOD detection~\cite{sedlmeier2019uncertainty, loquercio2020general}. We do not baseline against~\cite{lee2018simple} since this approach requires a $D_{out}$ dataset during the development phase.

\subsection{Datasets}
\label{sec:experiments:data}
We evaluate the proposed OOD detection method on two real-world automotive datasets: KITTI~\cite{Geiger2012CVPR} and nuScenes~\cite{nuscenes2019ceasar}. We extract \textit{Car}, \textit{Truck}, \textit{Cyclist}, and \textit{Pedestrian} object classes from the KITTI dataset represented as $192\times256\times3$ patches following the train/test split as suggested by \citet{nitsch2020learning}. Objects from the nuScenes dataset include ten different classes (\textit{Barrier}, \textit{Bicycle}, \textit{Bus}, \textit{Car}, \textit{Construction Vehicle}, \textit{Bike}, \textit{Officer}, \textit{Cone}, \textit{Trailer}, \textit{Truck}) with a patch size of $128\times128\times3$. We follow the train/validation split proposed by \citet{nuscenes2019ceasar}. In our experiments, the NuScenes validation set serves as the test set, because the test labels have not been released to the public. The KITTI and nuScenes datasets both serve as $D_{in}$, while the ImageNet~\cite{dengg2009imagenet} test set is chosen as $D_{out}$. The $D_{in}$ datasets are used for training whereas the $D_{out}$ dataset is purely used for OOD detection evaluation. The whole ImageNet test set is considered as $D_{out}$ although it might contain object classes from $D_{in}$ (e.g. cars). However, the amount of overlap between the classes in $D_{in}$ and $D_{out}$ is comparably small and it is the same for all methods during evaluation.

\subsection{Evaluation Metrics}
\label{sec:experiments:metrics}
We report the \textit{Detection Accuracy} and the \textit{Area Under Precision Recall (AUPR) curve} as done in related literature~\cite{lee2017training, lee2018simple}. Following the notation from \citet{lee2017training}, the \textit{Minimum Error Probability} $e$ is stated as:
\begingroup
\small
\begin{align}
\begin{split} \label{eq:tau}
    e = 1 - \min_\tau \big\lbrace &P\left( q(\mathbf{x}_{in}) \leq \tau \right) P\left(\mathbf{x}_{in}\right)  \\ 
    + &P\left( q(\mathbf{x}_{out}) > \tau \right) P\left(\mathbf{x}_{out}\right)\big\rbrace 
\end{split}
\end{align}
\endgroup
where $\mathbf{x}_{in}$ and $\mathbf{x}_{out}$ belong to $D_{in}$ and $D_{out}$, respectively, $P$ is the sample probability of x belonging to $D_{in}$ or $D_{out}$, $q$ is the confidence score of the evaluated metric (e.g. softmax, cosine similarity), and $\tau$ describes the best possible threshold between the confidence scores for $D_{in}$ and $D_{out}$. The \textit{Detection Accuracy} is defined as:
\begingroup
\small
\begin{align}
    \textit{Detection Accuracy} = 1 - e.
\end{align}
\endgroup
The \textit{Detection Accuracy} reflects the classification result only for the best possible threshold $\tau$ given knowledge of the confidence scores of the evaluated metric for $D_{in}$ and $D_{out}$. Therefore, it should be considered in combination with a threshold independent measure, such as \textit{Area Under Precision Recall (AUPR) curve}, as done in related literature~\cite{lee2017training, lee2018simple}. The AUPR is a suitable measure of the classification performance on class-imbalanced datasets, which is the case in our experiments. There are two subcategories in the AUPR metric: \textit{AUPR-in} ($D_{in}$ positive class) and \textit{AUPR-out} ($D_{out}$ positive class). The \textit{AUPR-in} metric is central to evaluating the performance of OOD detection as it characterizes how well $D_{in}$ is correctly identified during the classification task. For completeness, we report both \textit{AUPR-in} and \textit{AUPR-out}, as is done in~\cite{lee2017training, lee2018simple}.

\subsection{Quantitative Results}
\label{sec:experiments:quantitave}
\cref{tab:kitt_imagenet} and \cref{tab:nuscene_imagenet} show the results on the automotive datasets, KITTI ($D_{in}$) and nuScenes ($D_{in}$), with the ImageNet test set serving as $D_{out}$. The best result is highlighted in \textbf{bold} and the second best in \textit{italics}.
The results in \cref{tab:kitt_imagenet} and \cref{tab:nuscene_imagenet} confirm that the joint loss scheme is superior to the regular NN training for OOD detection. Furthermore, the cosine similarity is amongst the best results in all experiments with a clear advantage on the \textit{AUPR-in} metric. When considering the post hoc approaches to OOD detection on their own (CE term in \cref{eq:l_cls}), we found the cosine similarity to still be the superior choice for standard CE loss trained classifiers. $\chi^2$-based detection achieves comparable results but is nevertheless outperformed by the cosine similarity measure. We hypothesize that further improvements to covariance estimation may lead to better OOD detection performance with the $\chi^2$-based uncertainty.

For the KITTI experiments (\cref{tab:kitt_imagenet}), the cosine similarity metric achieves a $+5\%$ better \textit{Detection Accuracy} than softmax as well as a significant difference of $+10.91\%$ in the \textit{AUPR-in} metric. We hypothesize that its slightly lower performance on \textit{AUPR-out} compared to the baseline might be due to classes in the ImageNet test set overlapping with classes in $D_{in}$. The ImageNet test set does not provide labels, and therefore, it was not possible to easily filter out overlapping classes. Thus, we posit that our method is better able to discern the true OOD examples out of the imperfect $D_{out}$ dataset. For the NuScenes experiments (\cref{tab:nuscene_imagenet}), similar \textit{Detection Accuracy} is reported for the softmax and cosine similarity approaches. However, in the \textit{AUPR-in} metric, the cosine similarity metric clearly outperforms softmax with $+15\%$ improvement while sacrificing only $3\%$ on the \textit{AUPR-out}. Furthermore, the cosine similarity outperforms the MC dropout baseline on the \textit{Detection Accuracy} and \textit{AUPR-in} metrics in both datasets. Our approach has the advantage of being real-time capable, unlike MC dropout, which requires $T=100$ forward passes of the NN.

\begin{savenotes}
\begin{table}[]
\centering
\caption{\label{tab:kitt_imagenet} $D_{in}$ KITTI - $D_{out}$ ImageNet}
\begin{tabular}{|c|c|c|c|c|c|c|c|c|c|}
\hline
& Acc.\%$\uparrow$ & \makecell{Output} &  \makecell{Detection \\ acc. \%}$\uparrow$ & \makecell{AUPR \\in}$\uparrow$ &  \makecell{AUPR \\ out}$\uparrow$\\ \hline \hline
\multirow{4}{*}{\rotatebox[origin=c]{90}{CE loss}} &  \multirow{4}{*}{94.09} &  softmax  & 51.98 & 62.14 & 86.42\\ \cline{3-6}
& & MC dropout~\cite{blum2019fishyscapes} & 84.61 & 54.67 & 82.51\\ \cline{3-6}
& & $\chi^2$ (ours) & 86.49 & 76.12 & 65.20 \\ \cline{3-6}
& & CosSim (ours)  & 86.15 & \textit{81.34} & 80.08 \\ \hline\hline
\multirow{4}{*}{\rotatebox[origin=c]{90}{joint loss}} &  \multirow{4}{*}{93.30} &  softmax~\cite{lee2017training} & 84.61 & 78.95 & \textbf{94.67}\\ \cline{3-6}
& &  MI & 84.61 & 60.47 & \textit{90.56}\\ \cline{3-6}
& & $\chi^2$ (ours) & \textit{88.16} & 76.40 & 60.71\\ \cline{3-6}
& & CosSim (ours) & \textbf{89.61} & \textbf{89.86} & 90.25\\ \hline
\end{tabular} 
\vspace{-1mm}
\end{table}
\end{savenotes}

\begin{savenotes}
\begin{table}[]
\centering
\caption{\label{tab:nuscene_imagenet} $D_{in}$ nuScenes - $D_{out}$ ImageNet}
\begin{tabular}{|c|c|c|c|c|c|c|c|c|}
\hline
& Acc.\%$\uparrow$& \makecell{Output} & \makecell{Detection \\ acc. \%}$\uparrow$& \makecell{AUPR \\in}$\uparrow$ &  \makecell{AUPR \\ out}$\uparrow$ \\ \hline \hline
\multirow{4}{*}{\rotatebox[origin=c]{90}{CE loss}} &  \multirow{4}{*}{88.56} &  softmax & 83.85 & 64.52 & 81.67\\ \cline{3-6}
& &  MC dropout~\cite{blum2019fishyscapes} & 83.76 & 55.11 & 78.87\\ \cline{3-6} 
& &  $\chi^2$ (ours) & 83.80 & 65.92 & 64.40\\ \cline{3-6}
& &  CosSim  (ours) & 84.98 & \textit{88.82} & 82.84\\ \hline\hline
\multirow{4}{*}{\rotatebox[origin=c]{90}{joint loss}} & \multirow{4}{*}{88.85} & softmax~\cite{lee2017training} & \textbf{87.40} & 74.97 & \textbf{90.64}\\ \cline{3-6}
& &  MI & 83.79 & 56.88 & 82.12\\ \cline{3-6}
& & $\chi^2$  (ours) & 83.83 & 66.97 & 63.74\\ \cline{3-6}
& & CosSim  (ours)& \textit{87.13} & \textbf{91.51} & \textit{87.90}\\ \hline
\end{tabular} 
\vspace{-7mm}
\end{table}
\end{savenotes}

\subsection{Qualitative Results}
\label{sec:experiments:qualitave}
\begin{figure}
    \centering
     \begin{subfigure}{0.3\linewidth}
     \centering
     \includegraphics[width=\linewidth]{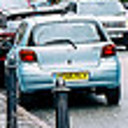}
     \caption{\tabular[t]{@{}l@{}}Basel.~\cite{lee2017training}:~\textcolor{correct_g}{$D_{in}$} \\ Ours:~\textcolor{correct_g}{$D_{in}$}\endtabular}\label{fig:qualitative_results:car_2}
     \end{subfigure}
    \begin{subfigure}{0.3\linewidth}
     \centering
     \includegraphics[width=\linewidth]{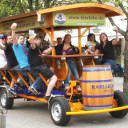}
     \caption{\tabular[t]{@{}l@{}}Basel.~\cite{lee2017training}:~\textcolor{correct_g}{$D_{out}$} \\ Ours:~\textcolor{correct_g}{$D_{out}$}\endtabular}\label{fig:qualitative_results:beer_bike}
     \end{subfigure}
     \begin{subfigure}{0.3\linewidth}
     \centering
     \includegraphics[width=\linewidth]{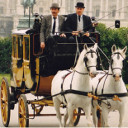}
     \caption{\tabular[t]{@{}l@{}}Basel.~\cite{lee2017training}:~\textcolor{fail_r}{$D_{in}$} \\ Ours:~\textcolor{correct_g}{$D_{out}$}\endtabular}\label{fig:qualitative_results:fiaker}
     \end{subfigure}
    \caption{The OOD detection results are computed using softmax (baseline~\cite{lee2017training}) and cosine similarity (ours) uncertainty measures from a network trained with the joint loss on the nuScenes dataset. Our approach classifies all objects \textcolor{correct_g}{correctly}, whereas the baseline \textcolor{fail_r}{fails} on the horse-drawn carriage example (\cref{fig:qualitative_results:fiaker}).}
  \label{fig:qual_result}
\vspace{-5mm}
\end{figure}
Qualitative OOD detection results are shown on a network trained with the joint loss using the nuScenes dataset ($D_{in}$). \cref{fig:qual_result} shows example road users that are not present in the nuScenes dataset alongside OOD detection results from the softmax~\cite{lee2017training} and cosine similarity (ours) detectors. The threshold used to classify a sample as $D_{in}$ or $D_{out}$ is the optimal $\tau$ evaluated in the \textit{Detection Accuracy} in \cref{tab:nuscene_imagenet}. The car in \cref{fig:qualitative_results:car_2} is not present in the nuScenes dataset ($D_{in}$) but visually belongs to the same distribution as the cars in the dataset, and is correctly classified as $D_{in}$. The example in \cref{fig:qualitative_results:beer_bike} is a \textit{beer bike} which can be seen in select German cities. This road user is not in the distribution of the nuScenes training data and both the softmax and the cosine similarity accurately classify it as $D_{out}$. \cref{fig:qualitative_results:fiaker} shows a \textit{Fiaker}, which is a special horse-drawn carriage unique to the city of Vienna and not present in the nuScenes dataset. The cosine similarity metric correctly detects it as $D_{out}$ whereas the softmax metric incorrectly classifies it as $D_{in}$. Although the baseline network was trained with auxiliary training techniques, softmax still assigns high probability to the horse-drawn carriage in \cref{fig:qualitative_results:fiaker}, classifying it as $D_{in}$ and mistaking it for a car with high certainty. This example highlights that the addition of post hoc statistics adds value to the auxiliary training methods in OOD detection.

\section{Conclusion and Future Work}
In this paper, we introduce an approach that effectively combines auxiliary training techniques and post hoc statistics to perform OOD detection. 
We investigate different post hoc metrics during inference to identify OOD samples. Our method with the cosine similarity metric achieves the best \textit{AUPR-in} performance on real-world datasets over the MC dropout~\cite{blum2019fishyscapes} and softmax~\cite{lee2017training} baseline approaches for OOD detection. In particular, the combination of generative networks during training and the post hoc computed class-conditioned Gaussian distributions achieves promising performance. We highlight that our proposed approach does not necessitate the definition and recording of an explicit $D_{out}$ dataset and only requires a single inference step of the NN.

As future work, our approach can be applied to the field of anomaly detection to recognize scenarios that are not yet present or underrepresented in $D_{in}$.
For example, a flat truck tire lying on the highway is a rarely recorded scenario but poses a high safety risk. Automated addition of these rare cases to the training database is important to make automotive perception robust and thereby ensure safe driving maneuvers. Although these scenarios might be manually annotated, they could easily be missed in the thousands of hours of data required for automotive validation processes.

\section*{Acknowledgement}
This work was funded in part by the Ford-Stanford Alliance and supported by the National Center of Competence in Research (NCCR) Robotics through the Swiss National Science Foundation.

% %%%%%%%%%%%%%%%%%%%%%%%%%%%%%%%%%%%%%%%%%%%%%%%%%%%%%%%%%%%%%%%%%%%%%%%%%%%%%%%%

\printbibliography
\end{document}